\documentclass{article}

\usepackage[T1]{fontenc}
\usepackage{float}
\floatstyle{plaintop}
\restylefloat{table}
\usepackage{amssymb,amsmath}
\usepackage{txfonts}
\usepackage{xspace}

\usepackage{microtype}
\usepackage{graphicx}
\usepackage{subfigure}

\usepackage{dirtytalk}

\usepackage{hyperref}

\usepackage[accepted]{icmlw2019generalization}

\icmltitlerunning{An Empirical Evaluation of Adversarial Robustness under Transfer Learning}

\begin{document}

\definecolor{burgundy}{rgb}{0.5, 0.0, 0.13}
\definecolor{armygreen}{rgb}{0.0, 0.26, 0.15}
\newcommand\boldgreen[1]{\textcolor{armygreen}{\textbf{#1}}}
\newcommand\boldblue[1]{\textcolor{blue}{\textbf{#1}}}
\newcommand\boldred[1]{\textcolor{burgundy}{\textbf{#1}}}

\twocolumn[
\icmltitle{An Empirical Evaluation of Adversarial Robustness under Transfer Learning}



\icmlsetsymbol{equal}{*}

\begin{icmlauthorlist}
\icmlauthor{Todor Davchev}{equal,ed}
\icmlauthor{Timos Korres}{equal,ed}
\icmlauthor{Stathi Fotiadis}{equal,ed}
\icmlauthor{Nick Antonopoulos}{equal,ed}
\icmlauthor{Subramanian Ramamoorthy}{ed}
\end{icmlauthorlist}

% \icmlaffiliation{to}{Department of Computation, University of Torontoland, Torontoland, Canada}
% \icmlaffiliation{goo}{Googol ShallowMind, New London, Michigan, USA}
\icmlaffiliation{ed}{School of Informatics, University of Edinburgh, Edinburgh, United Kingdom}

\icmlcorrespondingauthor{Todor Davchev}{t.b.davchev@ed.ac.uk}
% \icmlcorrespondingauthor{Eee Pppp}{ep@eden.co.uk}

\icmlkeywords{adversarial robustness, robust feature learning, transfer learning, robust optimisation}

\vskip 0.3in
]



\printAffiliationsAndNotice{\icmlEqualContribution} % otherwise use the standard text.

% This work evaluates adversarial robustness in the context of transfer learning from a source trained on CIFAR 100 to a target network trained on CIFAR 10. This allows us to identify transfer learning strategies under which adversarial defences are successfully retained, in addition to revealing potential vulnerabilities. We study the extent to which features learnt by a fast gradient sign method (FGSM) and its iterative alternative (PGD) can preserve their defence properties against black and white-box attacks under three different transfer learning strategies. We find that using PGD examples during training leads to more general robust features that are easier to transfer. Furthermore, under successful transfer, it achieves 5.2\% more accuracy against white-box PGD attacks than suitable baselines. In this paper, we study the effects of using robust optimisation in the source and target networks. Our empirical evaluation gives an insight on how how well adversarial robustness under transfer learning can generalise.

\begin{abstract}
In this work, we evaluate adversarial robustness in the context of transfer learning from a source trained on CIFAR 100 to a target network trained on CIFAR 10. Specifically, we study the effects of using robust optimisation in the source and target networks. This allows us to identify transfer learning strategies under which adversarial defences are successfully retained, in addition to revealing potential vulnerabilities. We study the extent to which features learnt by a fast gradient sign method (FGSM) and its iterative alternative (PGD) can preserve their defence properties against black and white-box attacks under three different transfer learning strategies. We find that using PGD examples during training on the source task leads to more general robust features that are easier to transfer. Furthermore, under successful transfer, it achieves 5.2\% more accuracy against white-box PGD attacks than suitable baselines. Overall, our empirical evaluations give insights on how well adversarial robustness under transfer learning can generalise.
\end{abstract}

\section{Introduction}
\label{sec:intro}
Machine learning models, in general, are known to be vulnerable to adversarial examples. For instance, certain imperceptible perturbations to the input can result in an incorrect classification \cite{szegedy2013intriguing, jia2017adversarial}. This should be concerning when such methods need to be deployed in safety-critical applications, such as autonomous vehicles or surgical robots \cite{warde201611, yuan2019adversarial}. A notable formulation of robustness against adversarial attacks is that of \citeauthor{madry2018towards} who formulate it as a "robust optimisation"-based problem: 
\[
 \min_{\theta}\mathbb{E}_{(x,y)\sim D}\max_{\epsilon} J(\theta,x+\epsilon,y) \\
 \quad   \text{ s.t} \quad
 \quad \epsilon \in \mathcal{S}
\]
% \vskip -2mm

Here, $J(\theta,x,y)$ is a loss specified in advance, where the tuple $(x,y)$ of the input and its label is sampled from some distribution $\mathbb{D}$. In this formulation, the adversary's task is to maximise the inner optimisation problem while the defender minimises the outer one. The adversary is specified through some threat model and is realised as a set of allowed perturbations $\mathcal{S}$. A defence can then be learned by augmenting the training data with adversarial examples \cite{42503}.

% Transfer learning is commonly used in deep learning for computer vision. However, adversarial attacks are a growing concern for learned models. To this end, little has been done towards assessing how well such defence mechanisms generalise to different tasks. We hypothesise that evaluating the efficacy of robust features under transfer can help us identify transfer learning strategies under which adversarial defences are successfully retained. Further, it can help us understand the ability of trained defence mechanisms to generalise against attacks that come from the same threat model. This can allow us to identify more informative ways to both defend and attack neural networks.

Transfer learning (TL) is commonly used deep learning technique. It has been showed to improve the performance as well as speed up training on variety of tasks \cite{pan2010surveytl}. To this end, little has been done towards assessing adversarial robustness under the scope of TL. We hypothesise that evaluating the efficacy of robust features under transfer can help us identify strategies under which adversarial defences are successfully retained. This can allow us to identify more informative ways to both defend and attack neural networks.
% The target result is to find more effective solutions in the context of transfer learning.
% To this end, little has been done towards assessing adversarial robustness under the scope of TL. We hypothesise that evaluating the efficacy of robust features under transfer can help us identify strategies under which adversarial defences are successfully retained. This can allow us to identify more informative ways to both defend and attack neural networks.
% The target result is to find more effective solutions in the context of transfer learning.

Our empirical evaluation indicates that adversarial robustness against black-box (BB) attacks transfers more consistently. In white-box (WB) scenarios, defence mechanisms benefit from using robust optimisation in both source and target. Further, adversarial training using PGD leads to learning more general robust features that can maintain their properties under transfer better than the alternative. With this, our empirical findings are the following: 

\begin{itemize}
    \item \textbf{Robustness}: We compared the level of transferred robustness between two tasks. PGD-based defences were easier to transfer than FGSM-based ones. Defending against simpler attacks sufficed from defending lower-level features only which are easier to transfer. %Change correlated features.

    \item \textbf{Generalisation}: We evaluate the ability to generalise against two threat models. We achieve \textbf{5.2\%} higher accuracy against WB PGD adversaries using robust weight initialisation as well as adversarial examples during training the target.
    
   \item \textbf{Performance}: We study the performance of different combinations of robust and clean optimisation routines. We visualise the results using normalised heatmaps and a complete table with accuracies.

\end{itemize}
\begin{figure}[t]
% \vskip 5mm
\begin{center}
\centerline{\includegraphics[width=0.4\textwidth]{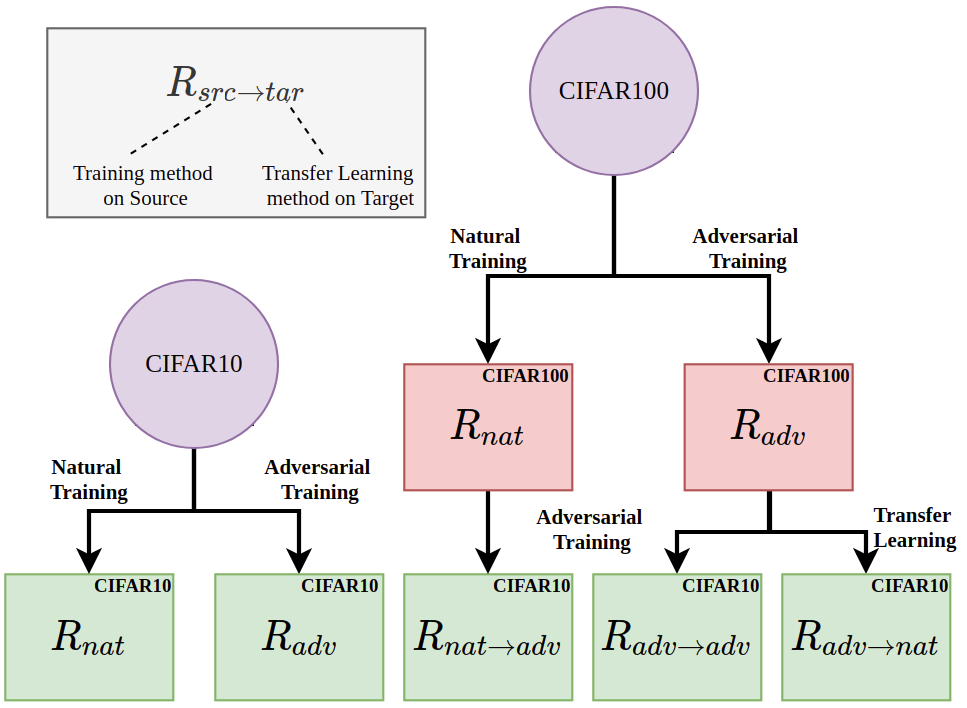}}
\caption{This figure represents a general outline of the acquired pipeline for evaluation. The subscripts follow the nomenclature for describing the network's way of training. Nat stands for clean training which means no adversarial examples were used whereas adv stands for adversarial training. For the networks obtained through transfer learning, we first mention the method of training its source (nat or adv) and then the method of transfer learning (nat or adv). In this context, the CIFAR100 networks (pink) are only an intermediary step on which we perform transfer learning.}
\label{fig:advTransLearning}
\end{center}
% \vskip -5mm
\end{figure}

\section{Methodology}

\subsection{Transfer learning}
\label{sec:Tranfer}
Transfer learning in CNNs can be achieved through retraining using various strategies \cite{yosinski2014transferable}. One can use the network as a feature extractor by freezing all the layers and only retraining the last one \cite{sharif2014cnn} or fine-tuning a larger part of the network \cite{oquab2014learning}.
% where fine-tuning the entire network essentially uses the learned features from source as an initialization point.
However, adversarial robustness may not necessarily be transferable in this process. So, we evaluate this property using the following three learning strategies: a) freeze all layers and retrain only the final layer, b) unfreeze only the last block of our network 
% and treat only the low-level of the trained source as a feature extractor;
and, c) retrain the whole network, essentially using the source as an initialisation strategy.

\subsection{Setup details}
In our experiments we used a Resnet56 network, denoted as $R$, which is an architecture specifically designed for the CIFAR dataset \cite{resnets}. Transfer learning is from \href{https://www.cs.toronto.edu/~kriz/cifar.html}{CIFAR100} to \href{https://www.cs.toronto.edu/~kriz/cifar.html}{CIFAR10}, where the images from both datasets where re-scaled to pixel values in [-1,1].

\subsection{Threat models}
As threat models, we use as adversaries the \textit{Fast Gradient Sign Method} (FGSM) \cite{43405} and the \textit{Projected Gradient Descent} (PGD) algorithm \cite{DBLP:journals/corr/KurakinGB16a} which is an iterative variant of FGSM.

FGSM creates an adversarial example $x'$ by following the gradient of the loss function with respect to the true label $y_{true}$. It then takes a single step $\epsilon$ towards that direction: 

% \vskip -4mm
\[ x' = x+ \epsilon\cdot sign  \nabla_x J(\theta, x, y) 
\]
% \vskip -1mm

PGD turns FGSM into an iterative attack, ensuring at each step that the adversarial example $x'$ is within the $l_{\infty}$ ball around $x$ with radius $\epsilon$:

% \vskip -4mm
\[
x ^k = Clip_{x,\epsilon}( x^{k-1} +  \alpha\cdot sign  \nabla_x J(\theta, x, y ))
\]

where  $k$ denotes the number of iterations and $\alpha$ the step taken at each iteration.

An intriguing property of FGSM is \textit{label leaking}, where the network achieves greater accuracy in the adversarial examples than with the clean ones \cite{DBLP:journals/corr/KurakinGB16a}. This happens most likely because FGSM produces a predictable perturbation which the network is able to identify. To avoid this effect, one  can replace the true label $y_{true}$ with the most likely label predicted by the model.

\textbf{Establishing the defence objective}

When performing adversarial training we modify the loss to be the average of the adversarial and clean loss, with equal weights  $L_{tot} =(L_{adv} + L_{clean})/2$. 
We weighted the two losses equally because we were aiming for both robustness and accuracy. Cross-entropy is used for the individual losses. 

Each batch had 200 examples, 100 clean images and 100 adversarial examples generated based on the most recent parameters of the network using the current clean examples. We used a standard $\epsilon = 0.0625$ which corresponds to pixel intensity of $\frac{8}{256}$. For the PGD adversary, we used 7 iterations and a learning rate $\alpha = \frac{\epsilon}{4}$ as was done by \citealt{madry2018towards}. We trained and used a DenseNet 121 to construct the BB attacks.

\section{Experimental Evaluation}
\label{sec:Experiments}
We evaluate the effect of using transfer learning in the process of building defence mechanisms against adversarial attacks. We perform a series of empirical evaluations and report our results in this section. 

Figure~\ref{fig:advTransLearning} introduces the general outline of the pipeline for evaluation. For brevity, we define $R_{src \rightarrow tar}$, where $ src, tar \in \{nat, adv\}$, as a network ($R$) learned with transfer learning. For example, $R_{nat \rightarrow adv}$ was trained with clean examples (nat) on the source domain and used both clean and adversarial examples (adv) during transfer. $R_{adv \rightarrow adv}$ was trained with both clean and adversarial examples from the same threat model in both source and target tasks. We evaluate the networks using the clean accuracy, as well as the accuracy against 4 adversaries, namely BB and WB attacks using FGSM and PGD.
% \subsection{Adversarial attacks}
% \textbf{White-box attacks}

% \textbf{Black-box attacks}
% To further investigate the robustness of these models, we tested them on black-box attacks. We crafted FGSM and PGD examples with a fixed $\epsilon = 0.0625$ using a naturally trained DenseNet on CIFAR10. In this scenario, it turned out that combining features from multiple domains by itself can increase robustness; except the case $R_{nat \rightarrow nat}$ all networks had increased robustness. Apart from that, black-box attacks illustrate similar behavior to white box, although they don't depend on local features of the network.

\subsection{Empirical Analysis of Transferring Robust Features}
\begin{figure}[h]
\centerline{\includegraphics[width=0.5\textwidth]{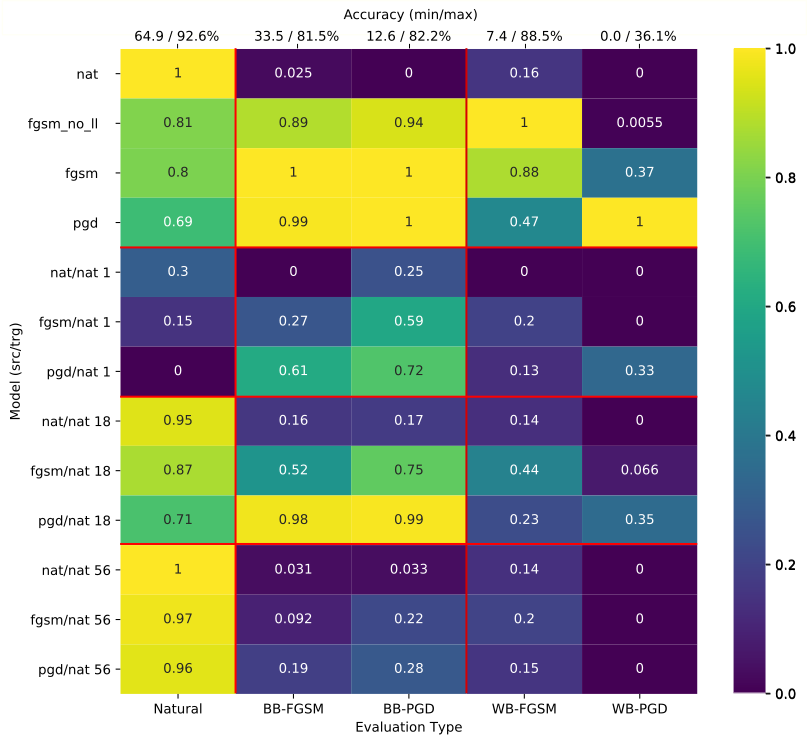}}
\caption{Heatmap comparison between the per column normalised results where 1 indicates the highest result and 0 the lowest. The table contains an empirical evaluation of the transferability of each defence routine for all considered strategies. Horizontal red lines separate transfer strategies where each model is named using the source/target type of training and the number describes the amount of unfrozen layers. No number indicates the model was trained without transfer and 56 indicates complete training using source's weights for initialisation.  '$no\_ll$' indicates no label leaking considered. Vertical red lines separate clean accuracy from the robustness achieved against WB and BB attacks. Percentage values on top represent the per column min/max accuracy. Overall using PGD in source results in higher ratios of transferred robustness compared to using FGSM. Keeping only lower-level features frozen leads to highest amounts of preserved robustness against BB attacks.}
\label{fig:TrasnferFeatures}
\end{figure}
Threat models have been shown to successfully attack both known and unknown architectures. Current defence systems have been targeting the actual attacks to a specific task and architecture. In practice, however, we would like to learn specific representations that can generalise to different settings. In this section, we evaluate the ability of defence systems to generalise to different tasks in the context of transfer learning. We examine the amount of transferable robustness from a source task to some target by training 3 independent neural networks on CIFAR100, namely: one that does not use adversarial examples during training and hence has no defence mechanism, one that uses examples generated using FGSM and a final network that uses examples generated using PGD. We report our results in Figure~\ref{fig:TrasnferFeatures}. 

The figure depicts the amount of transferred robustness across all strategies as a min-max normalised per attack heatmap. The first three rows show the robustness for the baseline networks that were trained with no transfer. Using PGD or FGSM samples during training performed equally well against BB attacks. In the former case, however, the resulted network is more robust against PGD-based WB attacks and less robust against FGSM-based ones. When training using FGSM generated samples we achieve the opposite results. These observations align with the ones made in \cite{madry2018towards} but we found FGSM to be more robust against PGD most likely because we consider label leaking. 

The next three blocks of rows (or 9 rows in total) report the results from transferring robustness using the three different strategies. Unlike the case in the absence of transfer, defence mechanisms developed using PGD are more likely to preserve the robustness of the learned features against BB attacks when used on the different task. However, both approaches seem to transfer proportionally the same amount of the achieved robustness against WB attacks. WB attacks are tailored for a specific architecture, hence directly transferring robustness was not expected to be as successful.

Overall, iterative learning results in more intricate and general features. However, the two tasks are distinct enough to not allow for the direct use of the learned features (see second block of results in Figure~\ref{fig:TrasnferFeatures}). Regardless, unfreezing the final block of ResNet56 resulted in an almost complete transfer of the defence against BB attacks. This itself suggests that robust low level features 
% from the lower levels of the network
are sufficient to defend just as well against such simpler attacks. 
% In the context of transfer, this enables the retraining of the final block of ResNet56 resulting in higher clean accuracy without any significant decrease in robustness against BB attacks. Defending lower level features seem to suffice in defending against BB attacks. 
Such features are easier to transfer too. Recent work, \cite{madry2019adversarialfeatures}, made similar observations and proposed a theoretical framework for studying such features. Unlike us, the authors do not focus on the transferability between tasks.

Evaluating against WB attacks, however, seems to be more successful at targeting aspects of the representation related to the higher levels of abstraction such as representations of the objects and sub-objects that are present in the input. Such attacks, however, have been shown to sometimes get stuck at local minimas resulting in weaker attacks \cite{DBLP:journals/corr/CarliniW16a}. Using robust features as initialisation does not seem to have as good of an effect when training the target network with clean examples only. That said, they can potentially allow for iterative methods to overcome the above limitations by ensuring better starting point for the optimisation procedure.

Finally, none of the reported methods fully matched or exceeded the performance of the baselines. In the next section we attempt to combine our findings with adversarial training applied on the target task as well.

% 
%  However, the defence mechanisms maintained their robustness against black-box attack. They preserved all of their robustness when freezing the low-level features in the context of training with PGD and most in the case of FGSM. This itself suggests that learning robust features in the initial layers  and retraining the final block of ResNet56 resulted in higher accuracy than its direct training alternatives suggesting the effect of pgd-trained defence mechanisms focus on the low-level features learned in a network only which can help improve the overall accuracy while preserving the robustness of the network. Further, it points to the different type of learned features for PGD and FGSM. Using robust features as initialization does not seem to have as good of an effect in direct transfer.
\begin{figure}[h]
\centerline{\includegraphics[width=0.5\textwidth]{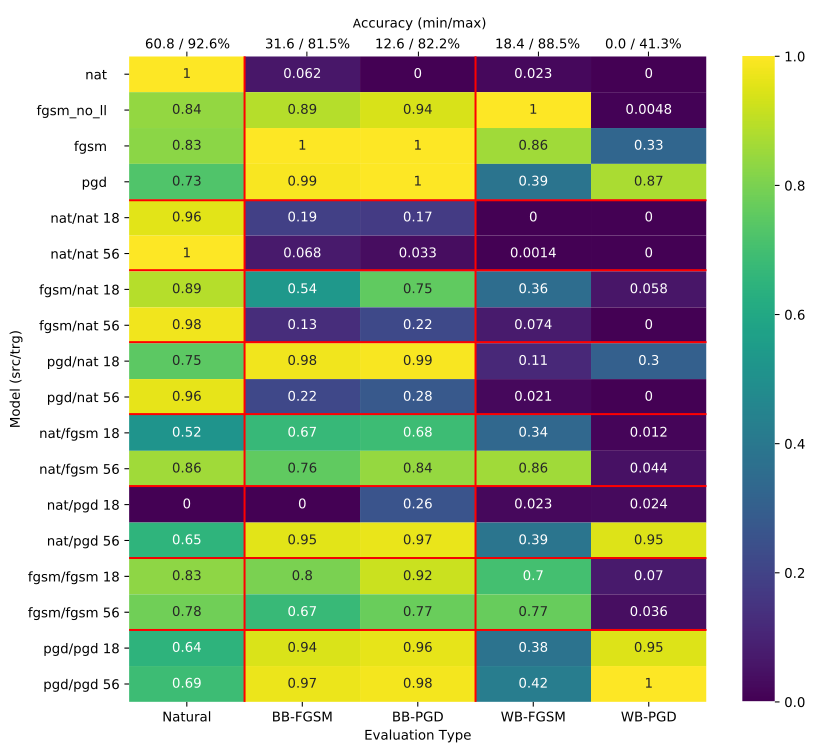}}
\caption{Heatmap comparison between the per column normalised results across different training routines. The table compares the use of adversarial examples in different combinations between source and target. Horizontal red lines separate different training routines. The numerical values within the heatmap are the per column normalised results. '$no\_ll$' indicates no label leaking taken into account. Vertical red lines separate clean accuracy from the robustness achieved against WB and BB attacks. Percentage values on top represent the per column min/max accuracy. Overall, good weight initialisation leads to improved performance of PGD-based defence mechanisms where robust initialisation is most effective.}
\label{fig:generalize}
\end{figure}

\subsection{Improving Defence Mechanisms with Transfer}

Successfully applying transfer learning has been shown to improve the performance as well as speed up training on a variety of tasks \cite{pan2010survey, yosinski2014transferable}. This suggests it can potentially enable building stronger, more general defence mechanisms as well as more complex attacks. We study the extent to which transfer learning can help us improve established defence mechanisms against adversarial attacks and the effects this has to clean accuracy. 

To this end, we compare the performance of an exhaustive list of models using adversarial attacks that follows the outline in Figure~\ref{fig:advTransLearning}.
Figure~\ref{fig:generalize} reports the performance of the best models per each of the 3 transfer learning strategies. Those omitted did not transfer robustness and are thus removed for brevity. The complete tables of results in \% and as a heatmap is provided in the Appendix. We use as baselines the non-transferred networks as well as networks that used transfer, but did not have any learned defence mechanisms. 

Using robust features as initialisation did not lead to positive results in the previous section. However, when combined with robust optimisation applied on the target network, it improved performance in the context of WB attacks while maintaining similar robustness as the baselines' against BB attacks. In fact, robust initialisation for $R_{pgd \rightarrow pgd}$ achieves 5.2\% accuracy improvement against WB PGD attacks, inline with recent observations about the properties of pre-training \cite{hendrycks2019using}. Training $R_{nat \rightarrow pgd}$ by unfreezing the last block of layers only did not result in successful transfer, even though we used adversarially perturbed examples during training. This is somewhat expected as we already saw that the lower level features are easier to attack and thus all attacks managed to exploit this. Finally, $R_{adv \rightarrow adv}$ seems to have a negative effect when using non-iterative methods.
% \begin{figure*}[t]
% \centerline{\includegraphics[width=1.0\textwidth]{robust_optm_empirical_strategic_normalized_across_all_redline.png}}
% \caption{Heatmap comparison between the normalized achieved accuracy across different training routines. The figure contains a comparison between using adversarial examples in source and target. The different strategies were split with horizontal red lines where strategy 0 indicates learning without transfer and the other 3 use transfer learning. The vertical lines separate natural accuracy from black-box (BB) and white-box (WB) robustness. Overall, unfreezing the final block of ResNet56 maintains the robst properties of the networks against BB attacks while initializing with robust weights and training the target network using adversarial examples achieved best results.}
% \label{fig:all_results}
% % \vskip -5mm
% \end{figure*} 
This itself again correlates with \cite{athalye2018obfuscated} and can be interpreted as ensuring that iterative attacks during training do not get stuck in local minima which itself ensures building a stronger defence.
% We must note that this could be caused from the learning rate annealing, since we initialised transfer with learning rate $0.1$.
Hypothetically, using a similar approach can lead to building stronger attacks too. Unfreezing the final block of ResNet gets close to the baseline results however requires less resources. Nevertheless, both networks obtain a lot worse clean accuracy.

\section{Conclusions and Future Work}
\label{sec:concl}

In this work, we investigate the use of transfer learning in the context of defending against adversarial perturbations. We showed that using FGSM and PGD during training results in different behaviour under transfer. PGD learns more general features that are easier to transfer to a different task. We found that lower level features by themselves play significant role in robustness against both WB and BB attacks and seem to be more transferable among tasks. Moreover, we showed that initialising with robust features can help improve the overall achieved robustness. When using PGD samples during re-training our analysis led to a 5.2\% robustness improvement against a WB PGD adversary for $R_{adv \rightarrow adv}$ compared to $R_{adv}$ and an overall stronger defence. A combination of freezing low-level features and training the final block of ResNet56 provides a good trade off that is both close to the best achieved  results while requiring a lot less training time. 

The reported results suggest that the current success against BB attacks can be achieved just by focusing on the lower level features of the network. On the other hand, WB attacks are able to target more complicated, higher level, "categorical" features, which makes it more challenging to defeat. Building attacks that can better exploit this observation could result in more challenging adversaries.
% \begin{itemize}
%     \item further investigation on a broader range of attacks and on different architectures;
%     \item evaluate performance on different tasks, such as control-based.
% \end{itemize}

In the future, we aim to further investigate the performance of defence mechanisms on a broader range of attacks and under transfer on different architectures. Further, we want to better understand the theoretical implications of the reported findings. Finally, we plan to extend the evaluation on control tasks in a simulated or a real-world setting.

\section*{Acknowledgements}
The authors would like to thank Antreas Antoniou for helpful discussions and technical advice, Michael Burke and Ben Krause for feedback on an early draft of the paper, and the anonymous reviewers for their comments. This work was supported in part by an EPSRC Industrial CASE award funded by Thales.

% \textbf{Do not} include acknowledgements in the initial version of
% the paper submitted for blind review.

% If a paper is accepted, the final camera-ready version can (and
% probably should) include acknowledgements. In this case, please
% place such acknowledgements in an unnumbered section at the
% end of the paper. Typically, this will include thanks to reviewers
% who gave useful comments, to colleagues who contributed to the ideas,
% and to funding agencies and corporate sponsors that provided financial

% % In the unusual situation where you want a paper to appear in the
% % references without citing it in the main text, use \nocite
% % \nocite{langley00}

\bibliography{example_paper}

\begin{thebibliography}{18}
\providecommand{\natexlab}[1]{#1}
\providecommand{\url}[1]{\texttt{#1}}
\expandafter\ifx\csname urlstyle\endcsname\relax
  \providecommand{\doi}[1]{doi: #1}\else
  \providecommand{\doi}{doi: \begingroup \urlstyle{rm}\Url}\fi

\bibitem[Andrew~Ilyas(2019)]{madry2019adversarialfeatures}
Andrew~Ilyas, Shibani~Santurkar, D. T. L. E. B. T. A.~M.
\newblock Adversarial examples are not bugs, they are features.
\newblock 2019.
\newblock URL \url{https://arxiv.org/abs/1905.02175}.

\bibitem[Athalye et~al.(2018)Athalye, Carlini, and
  Wagner]{athalye2018obfuscated}
Athalye, A., Carlini, N., and Wagner, D.
\newblock Obfuscated gradients give a false sense of security: Circumventing
  defenses to adversarial examples.
\newblock \emph{arXiv preprint arXiv:1802.00420}, 2018.

\bibitem[Carlini \& Wagner(2016)Carlini and
  Wagner]{DBLP:journals/corr/CarliniW16a}
Carlini, N. and Wagner, D.~A.
\newblock Towards evaluating the robustness of neural networks.
\newblock \emph{CoRR}, abs/1608.04644, 2016.
\newblock URL \url{http://arxiv.org/abs/1608.04644}.

\bibitem[Goodfellow et~al.(2015)Goodfellow, Shlens, and Szegedy]{43405}
Goodfellow, I., Shlens, J., and Szegedy, C.
\newblock Explaining and harnessing adversarial examples.
\newblock In \emph{International Conference on Learning Representations}, 2015.
\newblock URL \url{http://arxiv.org/abs/1412.6572}.

\bibitem[He et~al.(2015)He, Zhang, Ren, and Sun]{resnets}
He, K., Zhang, X., Ren, S., and Sun, J.
\newblock Deep residual learning for image recognition.
\newblock \emph{CoRR}, abs/1512.03385, 2015.
\newblock URL \url{http://arxiv.org/abs/1512.03385}.

\bibitem[Hendrycks et~al.(2019)Hendrycks, Lee, and Mazeika]{hendrycks2019using}
Hendrycks, D., Lee, K., and Mazeika, M.
\newblock Using pre-training can improve model robustness and uncertainty.
\newblock \emph{arXiv preprint arXiv:1901.09960}, 2019.

\bibitem[Jia \& Liang(2017)Jia and Liang]{jia2017adversarial}
Jia, R. and Liang, P.
\newblock Adversarial examples for evaluating reading comprehension systems.
\newblock \emph{arXiv preprint arXiv:1707.07328}, 2017.

\bibitem[Kurakin et~al.(2016)Kurakin, Goodfellow, and
  Bengio]{DBLP:journals/corr/KurakinGB16a}
Kurakin, A., Goodfellow, I.~J., and Bengio, S.
\newblock Adversarial machine learning at scale.
\newblock \emph{CoRR}, abs/1611.01236, 2016.
\newblock URL \url{http://arxiv.org/abs/1611.01236}.

\bibitem[Madry et~al.(2018)Madry, Makelov, Schmidt, Tsipras, and
  Vladu]{madry2018towards}
Madry, A., Makelov, A., Schmidt, L., Tsipras, D., and Vladu, A.
\newblock Towards deep learning models resistant to adversarial attacks.
\newblock In \emph{International Conference on Learning Representations}, 2018.
\newblock URL \url{https://openreview.net/forum?id=rJzIBfZAb}.

\bibitem[Oquab et~al.(2014)Oquab, Bottou, Laptev, and Sivic]{oquab2014learning}
Oquab, M., Bottou, L., Laptev, I., and Sivic, J.
\newblock Learning and transferring mid-level image representations using
  convolutional neural networks.
\newblock In \emph{Proceedings of the IEEE conference on computer vision and
  pattern recognition}, pp.\  1717--1724, 2014.

\bibitem[Pan \& Yang(2010{\natexlab{a}})Pan and Yang]{pan2010survey}
Pan, S.~J. and Yang, Q.
\newblock A survey on transfer learning.
\newblock \emph{IEEE Transactions on knowledge and data engineering},
  22\penalty0 (10):\penalty0 1345--1359, 2010{\natexlab{a}}.

\bibitem[Pan \& Yang(2010{\natexlab{b}})Pan and Yang]{pan2010surveytl}
Pan, S.~J. and Yang, Q.
\newblock A survey on transfer learning.
\newblock \emph{IEEE Transactions on knowledge and data engineering},
  22\penalty0 (10):\penalty0 1345--1359, 2010{\natexlab{b}}.

\bibitem[Sharif~Razavian et~al.(2014)Sharif~Razavian, Azizpour, Sullivan, and
  Carlsson]{sharif2014cnn}
Sharif~Razavian, A., Azizpour, H., Sullivan, J., and Carlsson, S.
\newblock Cnn features off-the-shelf: an astounding baseline for recognition.
\newblock In \emph{Proceedings of the IEEE conference on computer vision and
  pattern recognition workshops}, pp.\  806--813, 2014.

\bibitem[Szegedy et~al.(2013)Szegedy, Zaremba, Sutskever, Bruna, Erhan,
  Goodfellow, and Fergus]{szegedy2013intriguing}
Szegedy, C., Zaremba, W., Sutskever, I., Bruna, J., Erhan, D., Goodfellow, I.,
  and Fergus, R.
\newblock Intriguing properties of neural networks.
\newblock \emph{arXiv preprint arXiv:1312.6199}, 2013.

\bibitem[Szegedy et~al.(2014)Szegedy, Zaremba, Sutskever, Bruna, Erhan,
  Goodfellow, and Fergus]{42503}
Szegedy, C., Zaremba, W., Sutskever, I., Bruna, J., Erhan, D., Goodfellow, I.,
  and Fergus, R.
\newblock Intriguing properties of neural networks.
\newblock In \emph{International Conference on Learning Representations}, 2014.
\newblock URL \url{http://arxiv.org/abs/1312.6199}.

\bibitem[Warde-Farley \& Goodfellow(2016)Warde-Farley and
  Goodfellow]{warde201611}
Warde-Farley, D. and Goodfellow, I.
\newblock 11 adversarial perturbations of deep neural networks.
\newblock \emph{Perturbations, Optimization, and Statistics}, 311, 2016.

\bibitem[Yosinski et~al.(2014)Yosinski, Clune, Bengio, and
  Lipson]{yosinski2014transferable}
Yosinski, J., Clune, J., Bengio, Y., and Lipson, H.
\newblock How transferable are features in deep neural networks?
\newblock In \emph{Advances in neural information processing systems}, pp.\
  3320--3328, 2014.

\bibitem[Yuan et~al.(2019)Yuan, He, Zhu, and Li]{yuan2019adversarial}
Yuan, X., He, P., Zhu, Q., and Li, X.
\newblock Adversarial examples: Attacks and defenses for deep learning.
\newblock \emph{IEEE transactions on neural networks and learning systems},
  2019.

\end{thebibliography}
\bibliographystyle{icml2019}

\appendix
\begin{table*}[!htbp]
\begin{center}
\begin{tabular}{lrrrrr}
% \toprule
 Network & Natural & BB-FGSM  & BB-PGD & WB-FGSM & WB-PGD   \\
 \hline\hline
 $R_{nat}$  &   92.5\% &          34.7\% &         12.6\% &         20.0\% &         0.0\%  \\
 \hline
 $R_{nat  \rightarrow nat }$  1  &   73.1\% &          33.5\% &         30.2\% &          7.4\% &         0.0\% \\
$R_{nat  \rightarrow nat }$  18  &   91.2\% &          41.1\% &         24.6\%  &         18.4\% &         0.0\% \\
$R_{nat  \rightarrow nat }$  56  &   92.6\% &          35.0\% &         14.9\%  &         18.5\% &         0.0\% \\
\hline\hline
$R_{fgsm} $ &   87.1\% &        81.5\%  &     82.2\%       & 78.6\% &        13.5\%\\
$R_{fgsm\_no\_ll} $ &   87.4\% &        76.0\%  &     78.1\%       & 88.5\% &        0.2\%\\
 \hline
$R_{fgsm \rightarrow nat} $ 1  &   69.0\% &          46.3\% &         54.0\% &         23.5\% &         0.0\%  \\
$R_{fgsm \rightarrow nat} $ 18  &   89.1\% &          58.3\% &         65.1\% &         43.3\% &         2.4\% \\
$R_{fgsm \rightarrow nat} $ 56  &   91.9\% &          37.9\% &         27.9\% &         23.6\% &         0.0\% \\
\hline
$R_{nat  \rightarrow fgsm} $ 1  &   69.3\% &          40.9\% &         38.8\%&         13.1\% &         0.0\%  \\
$R_{nat  \rightarrow fgsm} $ 18  &   77.3\% &          65.0\% &         60.2\%  &         41.9\% &         0.5\%\\
$R_{nat  \rightarrow fgsm} $ 56  &   88.1\% &          69.3\% &         71.3\%  &         78.4\% &         1.8\%\\
 \hline
$R_{fgsm \rightarrow fgsm} $ 1  &   66.6\% &          56.2\% &         60.0\%   &     42.7\% &         0.2\%  \\
$R_{fgsm \rightarrow fgsm} $ 18  &   87.2\% &          71.4\% &         76.6\% &         67.5\% &         2.9\% \\
$R_{fgsm \rightarrow fgsm} $ 56  &   85.7\% &          65.2\% &         66.3\% &         72.3\% &         1.5\%  \\
 \hline\hline
 $R_{pgd}$  &   83.9\% &          81.1\% &        82.1\%&         45.4\% &        36.1\%  \\
 \hline
 $R_{pgd  \rightarrow nat }$ 1  &   64.9\% &          62.6\% &         63.0\% &         18.0\% &        12.0\% \\
$R_{pgd  \rightarrow nat }$ 18  &   84.6\% &          80.4\% &         81.6\%  &         26.2\% &        12.5\% \\
$R_{pgd  \rightarrow nat }$ 56  &   91.4\% &          42.4\% &         31.9\% &         19.9\% &         0.0\%  \\ \hline
$R_{nat  \rightarrow pgd }$  1  &   52.2\% &          30.6\% &         32.8\% &         13.2\% &         0.0\%  \\
$R_{nat  \rightarrow pgd }$  18  &   60.8\% &          31.6\% &         30.7\% &         20.0\% &         1.0\%  \\
$R_{nat  \rightarrow pgd }$  56  &   81.5\% &          79.2\% &         79.9\% &         45.6\% &        39.2\%  \\ \hline
$R_{pgd  \rightarrow pgd }$ 1  &   61.3\% &          59.6\% &         60.0\%&         25.6\% &        20.3\%  \\
$R_{pgd  \rightarrow pgd }$ 18  &   81.3\% &          78.6\% &         79.6\% &         45.1\% &        39.1\%  \\
$R_{pgd  \rightarrow pgd }$ 56  &   82.6\% &          80.1\% &         80.9\%&         47.7\% &        41.3\%  \\
\hline
% \bottomrule
\end{tabular}
\end{center}
\caption{The results on black-box (BB) and white-box (WB) attacks.  All networks were attacked with adversarial examples generated on the naturally trained DenseNet on CIFAR10. We achieve $5.2\%$ higher accuracy against WB PGD adversaries using robust weight initialisation as well as adversarial examples throughout the training of the target.}
\label{table:resultsinpercent}
\end{table*}
% \setcounter{page}{7}
% \begin{figure}
% \begin{center}
% \centerline{\includegraphics[width=0.9\textwidth]{Adversarial loss during training.pdf}}
% \caption{Training loss in adversarial attacks. FGSM training loss shows a sudden drop which is common in training with adversarial examples. The drop at 100 epochs in both methods is due to learning rate decay. PGD being a strongest attack does not reach 0 training loss.}
% \label{fig:advTransLearning}
% \end{center}
% \end{figure}

\begin{figure*}
\begin{center}
\centerline{\includegraphics[width=1\textwidth]{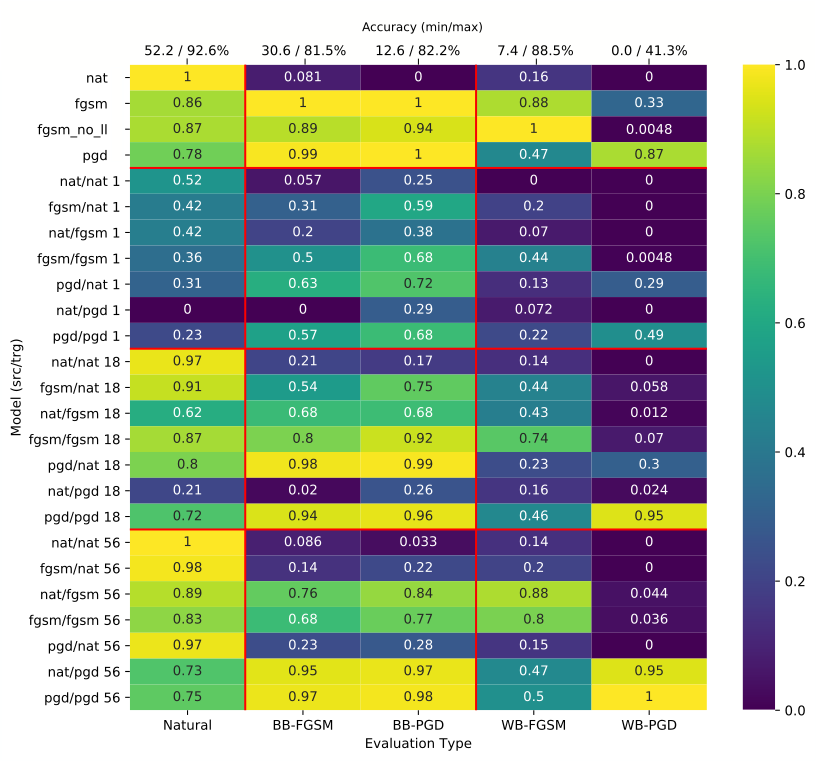}}
\label{fig:complete_plot}
\caption{The complete heatmap comparison between the per column normalised results. 1 indicates the highest result and 0 the lowest. The table contains an empirical evaluation of the transferability of each defence routine for all considered strategies. Horizontal red lines separate transfer strategies where each model is named using the source/target type of training and the number describes the amount of unfrozen layers. No number next to the model's name indicates the model was trained without transfer and 56 indicates complete training using source's weights for initialisation. '$fgsm\_no\_ll$' indicates an FGSM attack with no label leaking taken into account. Vertical red lines separate clean accuracy from the robustness achieved against WB and BB attacks. Percentage values on top represent the per column min/max accuracy. Overall using PGD in source results in higher ratios of transferred robustness compared to using FGSM. Transferring only lower-level features leads to highest amounts of preserved robustness against BB attacks.}
\end{center}
\end{figure*}

\begin{figure*}
\begin{center}
\centerline{\includegraphics[width=1\textwidth]{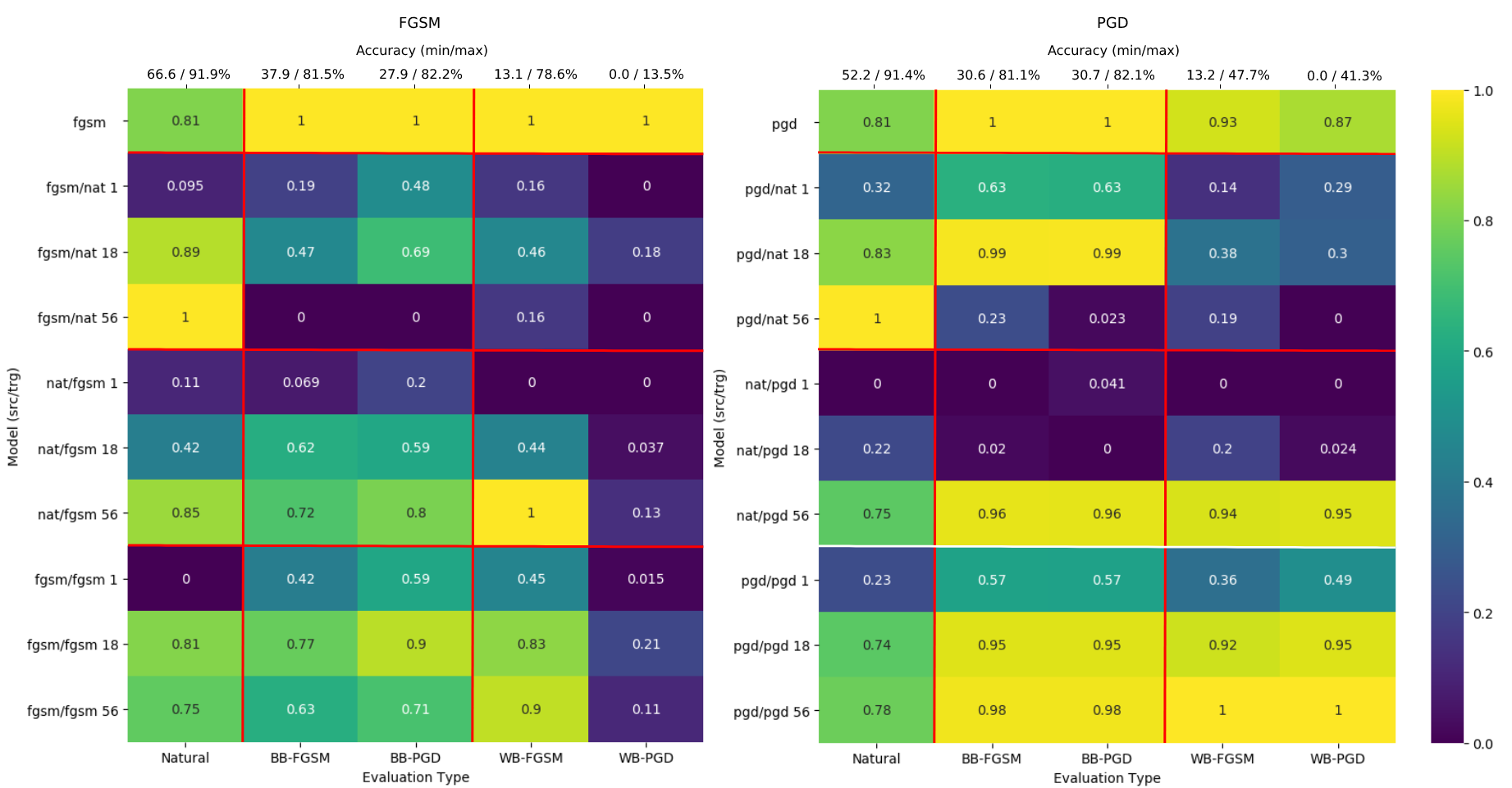}}
\label{fig:compare_pgd_fgsm}
\caption{Heatmap comparison between the per column normalised results in each table where 1 indicates the highest result and 0 the lowest. The table contains an empirical comparison between the transferability of FGSM against PGD for all considered strategies. Horizontal red lines separate the robust components across the different transfer strategies where each model is named using the source/target type of training and the number describes the amount of unfrozen layers. No number indicates the model was trained without transfer and 56 indicates complete training using source's weights for initialisation. Vertical red lines separate clean accuracy from the robustness achieved against WB and BB attacks. Percentage values on top represent the per column min/max accuracy. Overall using PGD defences is easier to transfer. Notice how the highest achieved results in the table reporting the use of FGSM is without transfer while this is not the case for the rightmost table reporting results obtained using PGD.}
\end{center}
\end{figure*}

\end{document}